\documentclass{article}
\usepackage{arxiv}
\usepackage{microtype}
\usepackage{graphicx}
\usepackage{subcaption}
\usepackage{booktabs} 
\usepackage{hyperref}

\usepackage{amsmath}
\usepackage{amssymb}
\usepackage{mathtools}
\usepackage{amsthm}

\usepackage{multirow}

\usepackage{enumitem} 
\usepackage{amsmath}  
\usepackage{graphicx} 
\usepackage[normalem]{ulem}
\useunder{\uline}{\ul}{}
\usepackage[capitalize,noabbrev]{cleveref}

\theoremstyle{plain}

\theoremstyle{definition}

\theoremstyle{remark}

\usepackage[textsize=tiny]{todonotes}

\title{A Knowledge-Informed Pretrained Model for Causal Discovery}
\author{
Wenbo Xu$^{1}$, Yue He$^{1}$, Yunhai Wang$^{1}$, Xingxuan Zhang$^{2}$, Kun Kuang$^{3}$, Yueguo Chen$^{1}$, Peng Cui$^{2}$ \\
\\
$^{1}$Renmin University of China, China \\
$^{2}$Tsinghua University, China \\
$^{3}$Zhejiang University, China \\
}
\begin{document}
\maketitle
\begingroup
\renewcommand{\thefootnote}{}
\footnotetext{Contact: Wenbo Xu (\url{feifeixwb@gmail.com}) and Yue He (\url{hy865865@gmail.com}).}
\addtocounter{footnote}{-1}
\endgroup
\begin{abstract}
Causal discovery has been widely studied, yet many existing methods rely on strong assumptions or fall into two extremes: either depending on costly interventional signals or partial ground truth as strong priors, or adopting purely data driven paradigms with limited guidance, which hinders practical deployment. 
Motivated by real-world scenarios where only coarse domain knowledge is available, we propose a knowledge-informed pretrained model for causal discovery that integrates weak prior knowledge as a principled middle ground. 
Our model adopts a dual source encoder-decoder architecture to process observational data in a knowledge-informed way.
We design a diverse pretraining dataset and a curriculum learning strategy that smoothly adapts the model to varying prior strengths across mechanisms, graph densities, and variable scales.
Extensive experiments on in-distribution, out-of distribution, and real-world datasets demonstrate consistent improvements over existing baselines, with strong robustness and practical applicability.

\end{abstract}

\section{Introduction}
Causal discovery aims to recover underlying dependencies from observational data and is fundamental to building reliable explanations, especially when randomized controlled trials (RCTs) are costly and ethically constrained.
It has broad impact in domains such as industrial process control~\cite{menegozzo2022bench}, biological networks~\cite{sachs2005causal}, environmental systems~\cite{stein2025causalrivers}, and policy evaluation\cite{athey2015machine}, where understanding causal mechanisms is often more valuable than prediction alone.
Structural Causal Models (SCMs)~\cite{pearl2009causality} provide a formal representation of causal relationships by encoding system mechanisms in a directed acyclic graph (DAG), where nodes denote variables and directed edges specify causal influence.
This formalism naturally motivates causal discovery algorithms that estimate a DAG (and associated mechanisms) from observational data, seeking structures that are consistent with the observed distribution.

Despite steady progress, causal discovery in real-world systems remains difficult due to heterogeneous mechanisms and noise. Existing methods rely on idealized assumptions and strong inductive biases that fail under mechanism mismatch~\cite{PC2014order,shimizu2006lingam,zheng2018notears}.
Some try to incorporate prior knowledge through rigid forms that require explicit edge specification~\cite{kalainathan2020cdt}.
Though achieving significant improvement, this rigidity limits flexibility and applicability.
Recent pretrained approaches improve inference efficiency but still leave room for improvement: they either rely on costly intervention signals~\cite{ke2022csiva,lorch2022Avici} or operate purely data-driven~\cite{meta}, leaving a gap between precise priors and pure observations.
\begin{figure}[t]
    \centering
    \includegraphics[width=0.8\linewidth]{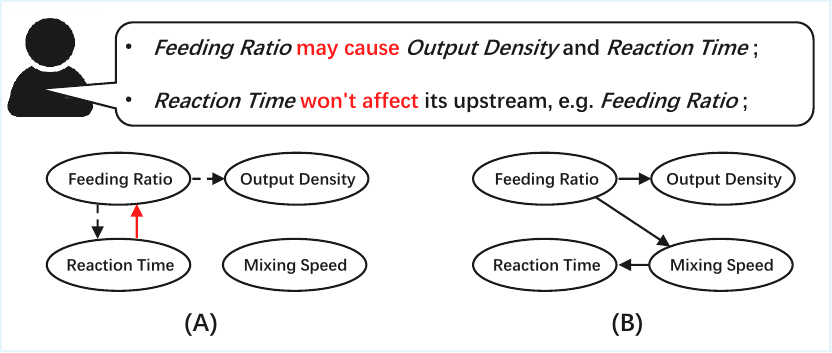}
    \caption{Simple causality in chemistry. (A) Expert knowledge is structured as graph. (B) The corresponding ground-truth causal graph.}
    \label{fig:examples_industry}
\end{figure}

Real-world knowledge is often available in a weaker form, while remaining informative. 
As illustrated in Fig.~\ref{fig:examples_industry}(A), an expert may distill experience into knowledge that both encourages (dashed lines) and rejects candidate relations.
Importantly, encouraged relations may be direct causal effects, e.g., \textit{Feeding Ratio} $\rightarrow$ \textit{Output Density}, or indirect one, e.g., \textit{Feeding Ratio} $\rightarrow$ \textit{Reaction Time} in Fig.~\ref{fig:examples_industry}(B). 
Both could provide heuristic guidance in causal discovery.
We will provide a more formalized definition in Section~\ref{knowledgeEncoding}.

To fully leverage such knowledge, we propose \textbf{\textit{Kode}}, a knowledge-informed pretrained model for causal discovery.
To the best of our knowledge, we propose the first pretrained model for causal discovery that jointly leverages observational data and prior knowledge.
Kode introduces a dual-source encoder-decoder structure that preserves the distinct semantics of data and knowledge while enabling cooperative inference.
To train such a model, we build a diverse synthetic training dataset spanning heterogeneous mechanisms, graph densities, and variable scales, together with a curriculum strategy that smoothly schedules prior strength.
Extensive evaluations including in-{/}out-of-distribution, and real-world dataset demonstrate consistent improvements over strong baselines, even without any knowledge input.
In summary, we have four contributions:
\begingroup 
\leftmargini=1.5em 
\begin{itemize} 
\item We introduce the first knowledge-informed pretrained model for causal discovery, establishing a new practical paradigm for real applications. 
\item We propose a dual-source encoder-decoder architecture that enables synergistic inference through semantic interaction between data and knowledge. 
\item We construct a diverse training corpus and implement a curriculum learning strategy to ensure seamless adaptation across varying data distributions and prior strengths.
\item We provide extensive evaluations demonstrating that our method consistently outperforms baselines across diverse scales, mechanisms, and sparsity levels.
\end{itemize} 
\endgroup

\section{Related Work}
Under the framework of Structural Causal Models (SCMs)~\cite{pearl2009causality}, a large body of work has been developed for causal discovery.
Existing approaches can be broadly categorized into three types~\cite{glymour2019review}.
Constraint-based approaches~\cite{spirtes1991PCalgorithm} infer causal relations via conditional independence tests.
Function-based methods achieve identifiability by imposing explicit assumptions on generating mechanisms~\cite{shimizu2006lingam,hoyer2008nonlinear}.
Score-based ones ~\cite{chickering2002ges} apply a specific score function to guide structure searching.
More recently, NOTEARS~\cite{zheng2018notears,zheng2020notears} introduces a differentiable score for DAG and has inspired a series of extensions~\cite{yu2019dag,ng2019gae,lachapelle2019gran,zhu2019rl}.
Despite their empirical success, these methods introduce distinct forms of inductive bias.
Traditional methods build on specific assumptions, while NOTEARS imposes additive equal-variance noise implicitly~\cite{reisach2021beware}.
To improve practical usability, prior knowledge is incorporated in implementation.
It's either used to reduce the search space~\cite{kalainathan2020cdt}, optimization space~\cite{sun2023nts, arthurmensch2023lbfgs} or as reward function~\cite{zhu2019rl,wang2021RLordering}.
And it's difficult to encode heuristic or uncertain domain knowledge, which is common in practice.

In recent years, Transformers have shown promising potential.
TabPFN \cite{hollmann2025pfn} demonstrates its sensitivity to causality.
Meanwhile, attention-based approaches for causal discovery have emerged.
They are pretrained on a synthetic dataset and infer via a single forward pass.
Representative methods include AVICI~\cite{lorch2022Avici}, CSIvA~\cite{ke2022csiva}, and BCNP~\cite{meta}, which share similar encoding paradigms but adopt distinct decoder designs with different modeling emphases.
However, these methods tend to fall into two extremes: either relies on interventional signals which are often costly in practice or follows a purely data-driven paradigm without any prior.
To bridge this gap, we propose a knowledge-informed pretrained model for causal discovery.

\section{Preliminary Study}
To develop a solution for the aforementioned difficulties, we first provide formal definitions for causal discovery and knowledge encoding with corresponding motivation.
\subsection{Problem of Causal Discovery}
Causal discovery is commonly framed as learning a directed acyclic graph (DAG) $\mathbf{G}=(\mathbf{V},\mathbf{E})$ that represents causal relations among variables $\mathbf{X}=\{X_1,\dots,X_d\}$, where each node corresponds to a variable and directed edges encode causal influences.
Under this formulation, causal discovery acts as a posterior estimator that infers the distribution over graph structures given observational data, i.e., $p(\mathbf{{G}}\mid \mathbf{X})$.

\subsection{Motivation}
As illustrated in Fig.~\ref{fig:examples_industry}, real-world knowledge is often available in a coarse-grained yet easily attainable form, while remaining informative.
Experts can readily provide high-level structural cues, such as procedural order or feasibility constraints, without specifying exact causal mechanisms. 
We abstract such information as an optional prior graph $G_P$.
When available, it provides auxiliary heuristic guidance for posterior inference in causal discovery $p(\mathbf{{G}}\mid \mathbf{X},G_P)$.
This view motivates the development of knowledge-informed models that can flexibly incorporate $G_P$ without over-reliance.

Unlike prior approaches that utilize partial ground-truth DAGs as hard constraints~\cite{zheng2020notears,kalainathan2020cdt}, we treat prior knowledge as soft, structure-aware signals that guide causal discovery without enforcing exact edges, avoiding direct edge supervision.
In many practical systems, like the industrial workflow in Fig.~\ref{fig:examples_industry}, experts may confidently rule out certain relations (e.g., downstream stages affecting upstream ones), identify relations that are plausibly influential, or leave others unspecified due to uncertainty.
These three cases can be categorized as \textit{may hold}, \textit{impossible}, and \textit{unknown}. 
Such coarse yet structure-aware signals preserve uncertainty inherent in human knowledge while still assisting the model in complex, large-scale graphs. 
Next, we formalize this representation of knowledge.

\subsection{Knowledge Encoding}
\label{knowledgeEncoding}
\begin{figure}[t]
  \centering
  \includegraphics[width=1\linewidth]{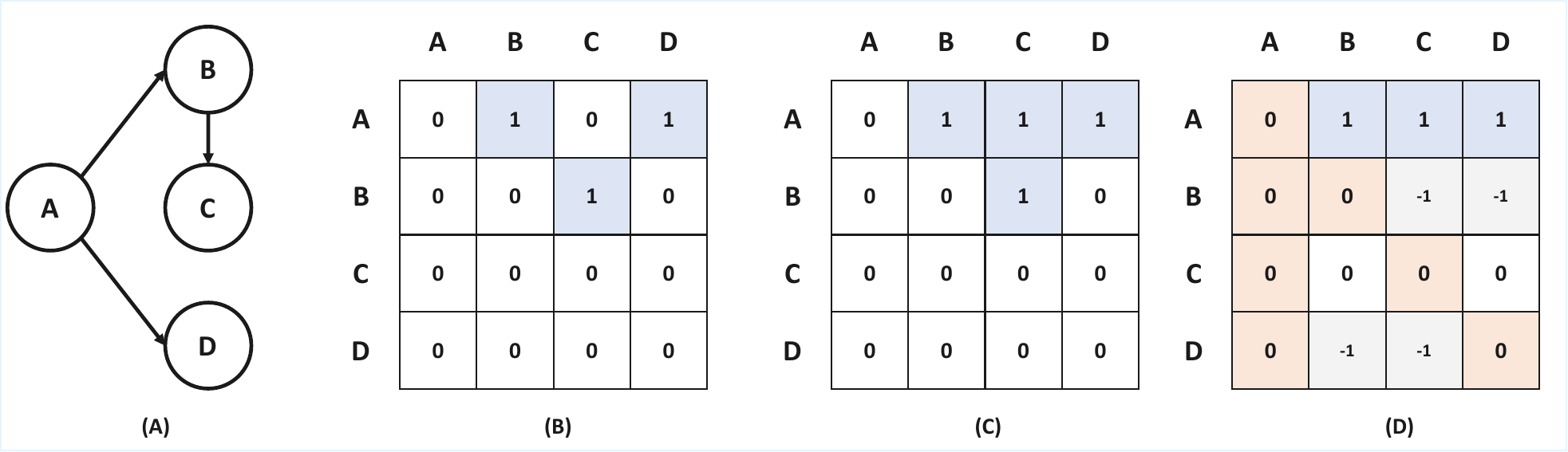}  \caption{An illustration of knowledge encoding.
(A) A symbolic form of Fig.~\ref{fig:examples_industry}.
(B) The corresponding adjacency matrix $\mathbf{A}$.
(C) The reachability matrix $\mathbf{R}$, which further encodes indirect relations (e.g., $R_{A,C}=1$).
(D) The resulting prior knowledge $G_P$, where infeasible relations are excluded and self-loops are removed.
}
  \label{pic:prior_encoding}
\end{figure}
Motivated by the three types of relations, we encode the knowledge structurally.
Specifically, we represent knowledge as a reachability matrix that captures potential causal influence without committing to exact edge existence.
Here, we define: 
$G_P = \mathbf{R} \in \{-1, 0, 1\}^{N \times N}$,
where $N$ denotes the number of variables.
Each entry $R_{ij}$ represents the belief about whether variable $i$ may causally affect variable $j$.
Specifically, we set $R_{ij}=0$ as \textit{impossible}, and set $R_{ij}=1$ to indicate that $i$ \textit{may cause} $j$, since connectivity within a reachability matrix does not explicitly distinguish between direct and indirect edges.
In real-world settings, prior knowledge is often incomplete.
Thus, we randomly assign entries to $-1$ to denote unknown relations.
Since causal graphs are assumed to be DAGs, we further enforce that $R_{ji}=0$ whenever $R_{ij}=1$,
and set all diagonal entries $R_{ii}=0$ to prevent self-loops.
Unknown masks are applied only to the remaining entries.
Notably, this structural representation provides scale-aware guidance even in the absence of positive entry $R_{ij}=1$, as the variable scale is inherently encoded within the knowledge matrix.

\begin{figure*}[t]
    \centering
    \includegraphics[width=1\linewidth]{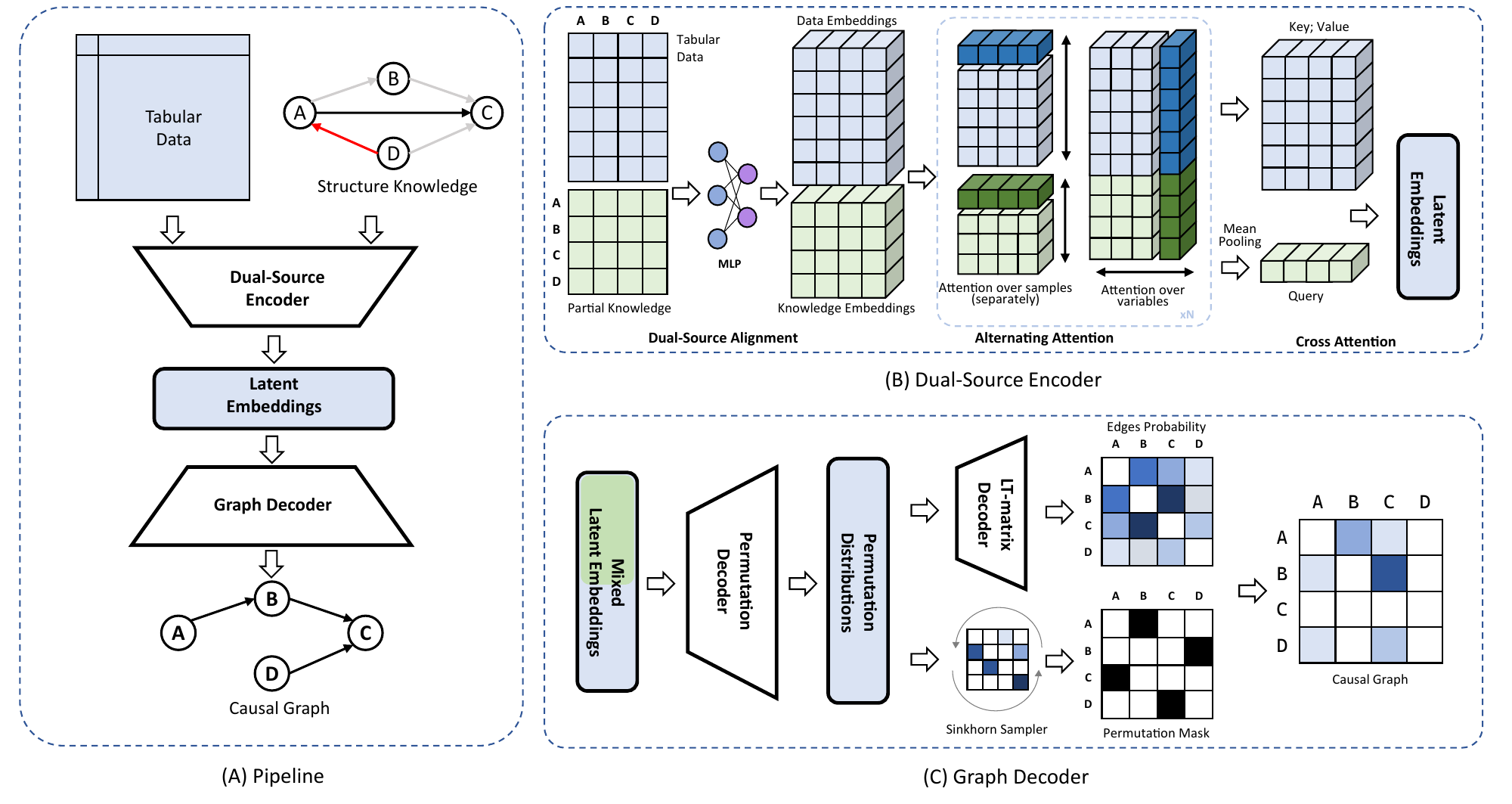}
    \caption{(A) The overall pipeline of Kode, illustrating the global framework that integrates raw data and prior knowledge to infer the causal graph; (B) The dual-source encoder, which achieves alignment and feature compression from hybrid inputs; (C) The graph decoder, responsible for mapping the latent representations back into the final causal structure.}
    \label{fig:pipeline}
\end{figure*}
\section{Model Architecture}
After encoding knowledge structurally, we then propose a knowledge-informed architecture with a dual-source encoder and a graph decoder, enabling an end-to-end pipeline that maps data to causal graphs in a knowledge-informed manner.
An illustration is shown in Fig.~\ref{fig:pipeline}.
\subsection{Dual-Source Encoder}
\subsubsection{Dual-Source Alignment}
With the prior knowledge encoded, we then align observational data and knowledge in a shared latent space.
Left part of Fig.~\ref{fig:pipeline}(B) illustrates details.
We denote the observations as: $\mathbf{X} \in \mathbb{R}^{S \times N}$
where $S$ and $N$ are the number of samples and variables.
Prior knowledge is encoded into $G_P \in \{-1, 0, 1\}^{N \times N}$ as mentioned previously.
Since they are heterogeneous semantically, we project them into a shared latent space, yielding:
$
\mathbf{E}_X \in \mathbb{R}^{S \times N \times d}, 
\mathbf{E}_P \in \mathbb{R}^{N \times N \times d}
$
where $d$ denotes the embedding dimension.
The two embeddings are then merged into a unified representation, where $S'=N+S$ and $[\cdot||\cdot]$ denotes dimension concatenation:
\begin{equation}
\mathbf{Z} = [\mathbf{E}_P \mathbin\Vert \mathbf{E}_X] \in \mathbb{R}^{S' \times N \times d}
\label{eq:input_emb}
\end{equation}

\subsubsection{Alternating Attention Mechanism}
Following previous work~\cite{hollmann2025pfn}, we adopt an alternating attention scheme.
Specifically, we design a multi-layer attention module where the attention axis alternates at each layer $\ell$.
For even-numbered layers, i.e., $\ell \equiv 0 \pmod{2}$, the model captures dependencies across samples.
Since $\mathbf{Z}$ consists of data and knowledge as Eq.~(\ref{eq:input_emb}), we derive: ${Z}^{(\ell)} = [\mathrm{Attention}_S(\mathbf{E}_P)\ || \ \mathrm{Attention}_S(\mathbf{E}_X)]$.
Importantly, although processed independently, $\mathbf{E}_{X}$ and $\mathbf{E}_{P}$ use the identical attention module, $\mathrm{Attention}_S(\cdot)$, because representations of knowledge and data could remain aligned in a shared relational space while preserving their distinct semantics.
For odd-numbered layers, i.e., $\ell \equiv 1 \pmod{2}$, the attention switches to the variable dimension.
In this stage, the model focuses on capturing latent dependency structures among variables within each sample.
Formally, we compute:
$
{Z}^{(\ell)}
= \mathrm{Attention}_N\!\left({Z}^{(\ell-1)}\right)
$.
In contrast to the sample-wise stage, $\mathbf{Z}$ is not separated, as attention across variables will not introduce semantic interference.
Alternately, the model is refined in the context of mutual perception between data and knowledge.
See \textit{Alternating Attention} in Fig.~\ref{fig:pipeline}(B) for illustration.

\subsubsection{Cross Attention Mechanism}
We introduce the cross attention to further integrate representations into a unified latent space.  
See \textit{Cross Attention} in Fig.~\ref{fig:pipeline}(B) for illustration.
Let \(L\) denote the final layer of the alternating attention, we have:
$
    E_P^{(L)} \in \mathbb{R}^{N \times N \times d}$ and $
    E_X^{(L)} \in \mathbb{R}^{S \times N \times d}
$.
We construct the \textit{Query} $\mathbf{q}$ with mean pooling:
$\mathbf{q} = \text{MeanPool}(E_P^{(L)}) \in \mathbb{R}^{N \times 1 \times d}$,
and let $E_X^{(L)}$ serve as the \textit{Key} $\mathbf{K}$ and \textit{Value} $\mathbf{V} $ in cross attention, given by:
$
\mathbf{H} = \text{Attention}(\mathbf{q}, \mathbf{K}, \mathbf{V}) \in \mathbb{R}^{N  \times d}
$.
Functionally, this design constructs a compact vector $\mathbf{H}$, allowing the model to retrieve and aggregate causal information.

\subsection{Graph Decoder}
\begin{table*}[tbp]
\centering
\caption{Training dataset configurations under different generative mechanisms.}
\resizebox{1\textwidth}{!}{%
\begin{tabular}{c c| c c c c c}
\hline
 &  & \multicolumn{2}{c}{\textbf{Linear}} 
 & \multicolumn{3}{c}{\textbf{Neural Networks}} \\

\textbf{Variables} &
\textbf{Edges} &
\multicolumn{2}{c}{$\text{noise} \sim \mathcal{N}(0,\,d^2)$} &
\multirow{2}{*}{$\text{Latent} \sim \mathcal{U}(-1,\,1)$} &
\multirow{2}{*}{$\text{Latent} \sim \Gamma(1,\,1)$} &
\multirow{1}{*}{$\text{Latent} \sim {N}(0,\,d^2)$} \\

 &  &
$d \sim \Gamma(2.5,\,2.5)$ &
$d \sim \Gamma(1,\,1)$ &
 & & $d \sim \Gamma(1,\,1)$\\ \hline

20 & $|\mathbf{E}| \in [20,\,60]$  & 20{,}000 & 30{,}000 & 16{,}667 & 16{,}667 & 16{,}667 \\
30 & $|\mathbf{E}| \in [30,\,90]$  & 20{,}000 & 30{,}000 & 16{,}667 & 16{,}667 & 16{,}667 \\
40 & $|\mathbf{E}| \in [40,\,120]$ & 20{,}000 & 30{,}000 & 16{,}667 & 16{,}667 & 16{,}667 \\ \hline
\end{tabular}%
}
\label{tab:data_generation}
\end{table*}
Following BCNP~\cite{meta}, our decoder samples DAGs as $\mathbf{G} = \mathbf{Q}\Lambda\mathbf{Q}^\top$, where $\mathbf{Q}$ is a permutation matrix and $\Lambda$ is a lower triangular one.
Before decoding, we regularize $\mathbf{H}$ by a weak additive knowledge signal with controlled interaction strength  $\epsilon \ll 1$, as
$\mathbf{H'} = \mathbf{H} + \epsilon \mathbf{q}$, which allows knowledge-informed inference to softly bias the representation while preserving data-driven signals, thereby guiding inference without disrupting the optimization landscape.

\subsubsection{Permutation Matrices}
As illustrated in Fig.~\ref{fig:pipeline}(C), we project $\mathbf{H'}\in \mathbb{R}^{N  \times d}$ into $\boldsymbol{\Theta} \in \mathbb{R}^{N \times N \times d}$ with attention-based \textit{Permutation Decoder}.
Then we adopt the Gumbel-Sinkhorn relaxation~\cite{mena2018sinkhorn} with the Hungarian hardening~\cite{charpentier2022hungarian}:
$
\tilde{\mathbf{Q}} \sim \operatorname{Sinkhorn}(\boldsymbol{\Theta}/\tau + \mathbf{G})
$, $
\mathbf{Q} = \operatorname{Hungarian}(\tilde{\mathbf{Q}}),
$
where $\tau$ and $\mathbf{G}$ control the process.
We use the straight-through estimator~\cite{bengio2013STE} to propagate gradients while keeping discreteness in the forward pass.

\subsubsection{Lower Triangle Decoder}
We parameterize edge existence with a lower triangular  matrix $\boldsymbol{\Phi}$, where the $\Lambda$ could be sampled. 
This naturally enforces acyclicity. 
To explicitly capture the semantic dependence and maintain representational consistency, we let $\mathbf{Q}$ and $\Lambda$ share parameters by feeding $\boldsymbol{\Theta}$ into the lower triangular decoder, as shown in Fig.~\ref{fig:pipeline}(C), where:
$\Phi_{ij} = LT\text{-}Decoder(\boldsymbol{\Theta})$
if $i < j$, and $\Phi_{ij}=0$ otherwise.
Consequently, the topological distribution of the causal graph is deterministically constructed as:
\begin{equation}
\hat{\mathbf{G}} = \mathbf{Q}\,\boldsymbol{\Phi}\,\mathbf{Q}^\top
\label{decoder}
\end{equation}

\section{Model Training}
We train the proposed model in a knowledge-informed manner that combines a diverse training dataset with a dedicated training strategy.
In addition, we design optimization objectives that promote stable learning. 
\subsection{Data Preparation}
\subsubsection{Generation Mechanism}
We construct a large-scale synthetic training dataset, which is diversified along three orthogonal dimensions: \emph{(i) Variable Scale}, \emph{(ii) Graph Sparsity}, and \emph{(iii) Generative Mechanism}.
This design exposes the model to a wide range of variations during training and is expected to promote robustness and generalization under complex and realistic scenarios.
As shown in Tab.~\ref{tab:data_generation}, the dataset contains $300{,}000$ distinct causal graphs with corresponding observational data.

For \emph{(i) Variable Scale}, we consider $N \in \{20, 30, 40\}$;
For \emph{(ii) Graph Sparsity} of each, we adopt the Erd\H{o}s--R\'enyi (ER) paradigm with a progressive setting from $\mathrm{ER}(1)$ to $\mathrm{ER}(3)$, where the number of edges $|\mathbf{E}|$ are from $[20, 60]$, $[30, 90]$, and $[40, 120]$, respectively.
As a result, the model is trained over a smooth distributional shift over structural complexity.
For the \emph{(iii) Generative Mechanism}, we consider linear additive noise model and MLPs for nonlinear model.

Specifically, for linear setting, we define a structural function:
$
    X_i = \sum_{j \in \mathrm{pa}(i)} w_{ij} X_j + \epsilon_i,
$
where the weight $w_{ij}$ is sampled from a uniform distribution $w_{ij} \sim \mathcal{U}(-10, 10)$, and heteroscedastic noise $\epsilon_i$ is sampled as:
$
\epsilon_i \sim \mathcal{N}(0, \sigma_i^2),  \sigma_i \sim \Gamma(\cdot, \cdot).
$
In our implementation, we consider two distinct gamma distributions $\Gamma(\cdot, \cdot)$, namely $\Gamma(2.5, 2.5)$ and $\Gamma(1, 1)$.
This design differs from the standard homoscedastic assumption of identifiability \cite{zheng2018notears}, thereby providing a more challenging yet realistic training signal.

For the nonlinear setting, each variable is generated by a multi-layer perceptron (MLP), taking both its parent variables and an additional latent variable as input. 
Formally, we define:
$X_i = \mathbf{MLP}_i(\mathbf{x}_{\mathrm{pa}(i)}, z_i)$,
where the input is formed as $(\mathbf{x}_{\mathrm{pa}(i)} \oplus z_i)$, thereby introducing a non-additive stochasticity.
Each edge of generation function is independently initialized as a three-layer MLP with hidden width 32.
For each layer, we establish: 
$Output_l = Layer_{l} (\phi (Output_{l-1}))$ when $l = 2,\,3$
and
$Output_l = Layer_l (\mathbf{x}_{\mathrm{pa}(i)} \oplus z_i)$ when $l = 1$, 
where $\phi$ denotes the LeakyReLU activation.

To balance coverage across graph sizes, densities, and mechanisms, we adjust the samples, see Tab.~\ref{tab:data_generation} for details.
This yields a diverse training distribution that promotes robustness across scale, sparsity, and generative variations.

\subsubsection{Sampling of Structural Priors}
\begin{table*}[b]
\centering
\caption{Sampling probabilities for different prior types.}
{
\begin{tabular}{cccc}
\hline
\multicolumn{2}{c}{\textbf{Prior Type}}        &  & \textbf{Probability} \\ \hline
\multirow{2}{*}{Reachable Matrix} & (0, 1, -1) &  & 0.45                 \\
                                  & (0, -1)    &  & 0.45                 \\ \cline{1-2}
\multirow{2}{*}{Ground Truth}     & (0, 1, -1) &  & 0.05                 \\
                                  & (0, -1)    &  & 0.05                 \\ \hline
\end{tabular}}
\label{tab:prior_distribute}
\end{table*}

To simulate uncertainty in prior knowledge, we randomly mask the reachability matrix $G_P$.
Concretely, we sample a binary mask $\mathbf{M} \in \{0,1\}^{N \times N}$ from the Bernoulli distribution:
$
M_{ij} \sim \mathrm{Bernoulli}(\rho),
$
with preservation rate $\rho \in [0,1]$.
For each entry $(i,j)$, the structured prior knowledge is retained if $M_{ij} = 1$; and set to $-1$ (unknown) otherwise.
During training, we augment knowledge of reachability matrix with a small fraction of ground-truth graphs as anchors to guide and stabilize the learning process.
We adopt $G_P\in\{1,0,-1\} ^{N \times N}$ and $G_P\in\{0,-1\}^{N \times N}$ as different prior strength.
Tab.~\ref{tab:prior_distribute} provides a detailed description.

To facilitate optimization, unreachable and padding edges are assigned $10^{-10}$ rather than $0$ to preserve weak signal activity for gradient flow.
In contrast, diagonal entries are fixed to zero to strictly prohibit self-loops.

\subsubsection{Max-Variable Padding Strategy}
To accommodate different graph scales, we adopt zero-padding to fix maximum size.
In our knowledge-informed setting, padding in the observations $\mathbf{X}$ can be viewed as isolated points, while padding in the knowledge $G_P$ indicates that such points have neither parents nor children topologically. 
Formally, for any $N < N_{\max}$, we have
$
\tilde{\mathbf{X}} = [\mathbf{X}, \mathbf{0}] \in \mathbb{R}^{S \times N_{\max}}    
$
, and:
$$
\tilde{G_P} =
\begin{bmatrix}
G_P & \mathbf{0} \\
\mathbf{0} & \mathbf{0}
\end{bmatrix}
\in \{0,1,-1\}^{N_{\max} \times N_{\max}} 
$$

Similarly, we use $10^{-10}$ instead of 0 to preserve gradients for padding.
To handle dynamic graph scales, there are several different size-agnostic strategies~\cite{ke2022csiva,lorch2022Avici,meta}, but may induce a risk of scale-dependent variations in attention allocation~\cite{qin-etal-2022-devil}, especially in knowledge-informed settings due to scale mismatches between data and knowledge but with shared parameters.
In contrast, our padding strategy with small constant promotes size invariance, while enabling more stable optimization.
More importantly, when combined with structural knowledge, the padded representations explicitly encode the absence of interactions for non-existent variables, allowing the model to flexibly perceive effective graph size and adapt its causal reasoning across different variable scales.
\subsection{Curriculum Learning Strategy}
We adopt a curriculum learning strategy~\cite{bengio2009curriculum} to jointly schedule the proportion and strength of sparsity, enabling a stable and progressive training process.
From an optimization perspective, curriculum learning organizes training signals from easy to hard, allowing the model to first adapt under mild conditions before gradually facing more restrictive constraints.
Such a gradual progression stabilizes optimization and improves convergence by avoiding abrupt changes in the learning landscape.

In the context of the knowledge-informed model, a progressive training is particularly important. 
Sparse knowledge ensures robust optimization, but limited information is always hard to exploit at an early stage, whereas complete and dense knowledge may dominate data-driven learning and cause the model to depend excessively on knowledge.
To mitigate these issues, we adopt a staged training scheme that exposes the model to knowledge with diverse sparsity levels before progressively emphasizing more restrictive sparse knowledge.
This progression incrementally bolsters knowledge-informed inference while ensuring robustness through the synergy between data and prior knowledge.
As a result, the training process can be characterized by two key factors: the proportion of sparse knowledge and the strength of sparsity constraints.
Formally, at epoch $e$, the sparsity ratio $\rho_e$ is sampled from a mixture distribution:
\begin{equation}
\rho_e \sim
\begin{cases}
\mathrm{Uniform}(0,1), & \text{with probability } 1 - \pi(e), \\
\mathrm{Uniform}(0, \rho_{\max}(e)), & \text{with probability } \pi(e),
\end{cases}
\end{equation}
where $\pi(e)$ is a monotonically increasing function controlling the probability of sparse sampling.
Moreover, the upper bound of sparsity follows a linear annealing schedule:
$
\rho_{\max}(e) = \max(\tau_0 - \gamma e, \rho_{\min}),
$
where $\tau_0=0.5$ is the initial threshold, $\gamma=0.01$ is the annealing rate, and $\rho_{\min}=0.2$ prevents over-constraining the model.

\begin{table*}[b]
\centering
\caption{Illustration how the sampling proportions and the sparsity threshold evolve over training epochs.}
\begin{tabular}{ccccc}
\hline
\multicolumn{1}{c}{\textbf{Epochs}} &  & \textbf{Proportion} & \multicolumn{1}{l}{} & \textbf{Sparse Threshold} \\ \hline
{[}0, 5)    &  & No Sparse &  & /             \\
{[}5, 8)    &  & 8 : 2 &  & (0.34, 0.40{]} \\
{[}8, 10)   &  & 6 : 4 &  & (0.30, 0.34{]}\\
{[}10, 15)  &  & 4 : 6 &  & (0.20, 0.30{]}  \\
{[}15, 50{]} &  & 2 : 8 &  & 0.20           \\ \hline
\end{tabular}
\end{table*}

To this end, we implement a curriculum learning strategy in knowledge-informed training, where sparser knowledge is sampled more frequently and enforced more strictly over epochs.
As a result, the model transitions smoothly toward a more sparse and challenging hypothesis space, thereby mitigating the impact of abrupt distribution shifts.

\subsection{Optimization Objective}
\subsubsection{Similarity Constraint}
To enforce structural consistency between the representation and output spaces, we introduce a similarity-based constraint. The key intuition is that relative relationships among samples in the embedding space should be preserved in the outputs. We compute the embedding similarity matrix using cosine similarity,
$S^{\text{emb}}_{ij} = \mathrm{sim}_{\cos}({H}_i, {H}_j),
$
and define the corresponding output similarity via pairwise Euclidean distances,
$
S^{\text{out}}_{ij} = {1}/({1 + \|\hat{\mathbf{y}}_i - \hat{\mathbf{y}}_j\|_2}).
$
Both similarity matrices are standardized to remove scale effects. The objective minimizes their mean squared discrepancy:
$\mathcal{L}_{\text{sim}} = \| S^{\text{emb}} - S^{\text{out}} \|_2^2.
$
This constraint promotes structure-aware representations by aligning  across the two spaces.

\subsubsection{Graph Likelihood}
Following the paradigm of BCNP~\cite{meta}, we characterize the existence of each edge as an independent Bernoulli random variable according to Eq.~(\ref{decoder}). 
By marginalizing over multiple Monte Carlo samples, the training objective is formulated as the minimization of the negative log-likelihood (NLL):
\begin{equation}
\begin{aligned}
\mathcal{L}_{\text{graph}}
=
-\log p(\mathbf{G})
=
-\log \left( \frac{1}{K} \sum_{k=1}^K p(\mathbf{G} \mid \hat{\mathbf{G}}^{(k)}) \right)
\end{aligned}
\end{equation}
To enforce the structure-aware optimization, we minimize the combination of objective:
\begin{equation}
\mathcal{L}
=
\mathcal{L}_{\text{graph}}
+
\alpha \mathcal{L}_{\text{sim}}
\end{equation}
where $\alpha$ controls the strength of the similarity regularizer.

\section{Experiments}
\begin{figure*}[t]
    \centering
    \includegraphics[width=1\linewidth]{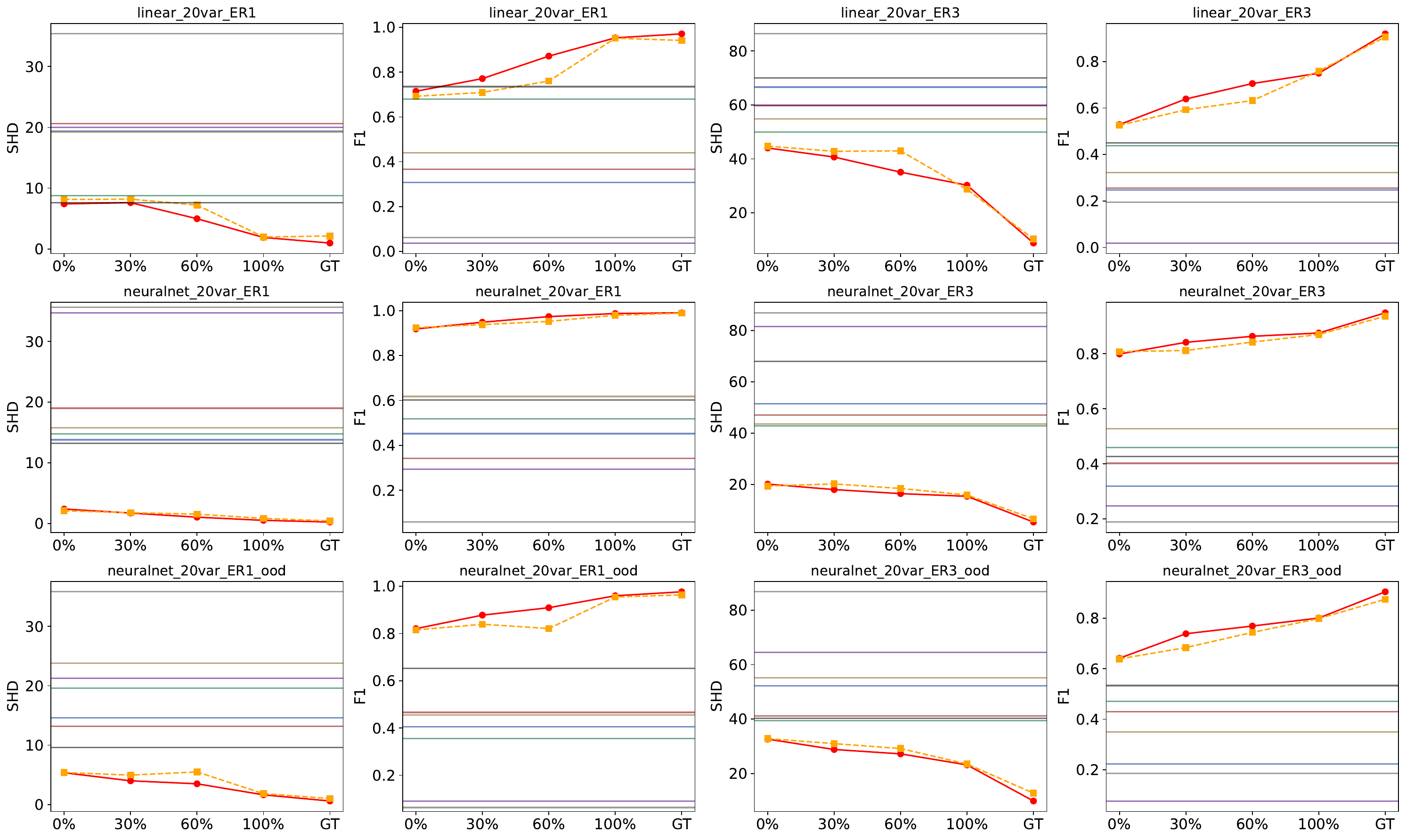}
    \includegraphics[width=0.5\linewidth]{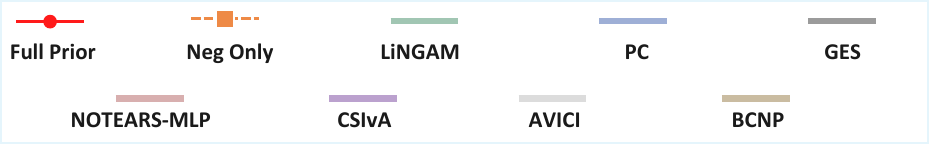}
    \caption{Results on 20 nodes, covering i.d. linear cases and both i.d. \& o.o.d nonlinear mechanisms.}
    \label{fig:result_main}
\end{figure*}
In this section, we briefly introduce the baselines and evaluation metrics, and report experimental results on both synthetic and real-world datasets. 

\subsection{Setup}
\label{setup}
\subsubsection{Baseline}
We compare our method with a comprehensive set of representative baselines, including function-based, score-based, and constraint-based approaches, as well as other pretrained models.
\textbf{LiNGAM}\cite{shimizu2006lingam}: a function-based method that assumes linear non-Gaussian noise;
\textbf{PC}\cite{PC2014order}: a constraint-based algorithm, searches structure via CI tests;
\textbf{GES}\cite{chickering2002ges}: a score-based framework, searches structure greedily;
\textbf{NOTEARS-MLP}\cite{zheng2018notears,zheng2020notears}: a seminal framework to learn a differentiable DAG score;
\textbf{CISvA}\cite{ke2022csiva}: the first pretrained model for causal discovery;
\textbf{AVICI}\cite{lorch2022Avici}: improving CISvA by decoding causal graph edge by edge;
\textbf{BCNP}\cite{meta}: the first fully data-driven meta-learner for causal discovery.

\subsubsection{Metrics}
We evaluate the estimated causal graphs with two metrics.
\textbf{SHD}: the minimum edit distance to the ground-truth graph.
The lower, the better.
\textbf{F1 score}: the harmonic mean of precision and recall. 
The higher, the better.

\subsection{Synthetic Data}
To evaluate our knowledge-informed model, we adopt two types of knowledge: a standard prior encoded as $\{-1,0,1\}$ and a weaker prior containing only negative one, encoded as $\{-1,0\}$. 
Both are derived from reachability matrix of the ground-truth.
Empirically, most sampled edges represent indirect causal effects (i.e., ancestors rather than parents), thereby mitigating the risk of direct edge supervision.

These settings are denoted as \textit{Full Prior} and \textit{Neg Only} in Fig.~\ref{fig:result_main}. 
Edges are randomly masked according to a retention ratio (x-axis), where 0\% corresponds to purely observational data and GT represents the special case where the full ground-truth graph is provided as prior knowledge. 
Performance under varying prior strengths is shown using line plots, with baseline results indicated by horizontal bars. 
Each subplot reports both SHD and F1 for the same mechanism under varying prior strengths.

\begin{figure}
    \centering
    \includegraphics[width=\linewidth]{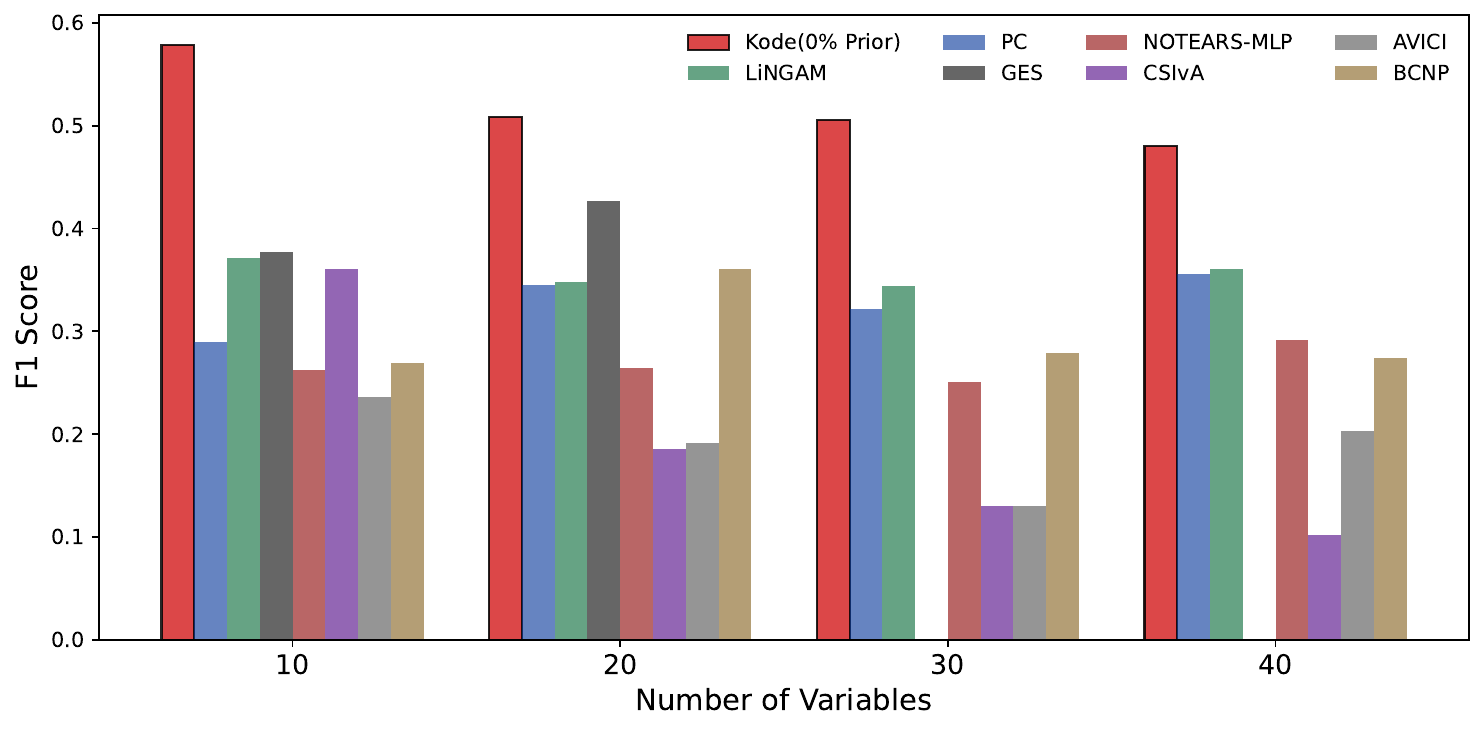}
    \caption{Performance on Gaussian Process ER(3) across scales: \textit{\textbf{Kode}} maintains a clear advantage; GES ($4^{th}$ bar) collapses owing to excessive density at $N=30$ and $40$, so its results are omitted.}
    \label{fig:varibles}
\end{figure}

We evaluate our model along four dimensions:
(i) \textit{Graph Densities},
(ii) \textit{Mechanisms},
(iii) \textit{Scales}, and
(iv) \textit{Out-Of-Distribution (o.o.d.) Generalization}.
For (i), (ii), and (iv), we evaluate performance under both linear and nonlinear processes across in-distribution and o.o.d. scenarios, with results for ER(1) and ER(3) graphs reported in Fig.~\ref{fig:result_main}. 
For (iii), we illustrate the model's scalability and performance across varying node scales $N \in \{10, 20, 30, 40\}$ in Fig.~\ref{fig:varibles}.
The first two rows of subplots evaluate performance on test data following the training distribution.
The third row, denoted as NN$_{\text{o.o.d}}$, introduces a severe distribution shift by increasing the network depth to 5 layers, replacing ReLU with sigmoid activations, sampling $\text{Latent}\sim \Gamma(2.5,\,2.5)$ and initializing weights randomly from $\text{W}\sim\mathcal{U}(-50,50)$.

For each experimental setting, we run 100 independent trials and reported the averaged performance to ensure robustness. 
As the proportion of knowledge increases, the performance of our method improves consistently; meanwhile, gains from negative knowledge are predictably less pronounced than from full knowledge.
While baselines like GES excel under linear mechanisms due to its regression-based likelihood scoring, our method outperforms all baselines even without prior knowledge, indicating strong data-driven capability.
These results confirm that our model effectively internalizes diverse mechanisms and scales, enabling robust inference and zero-shot generalization.

\subsection{Real-world Data}
To demonstrate applicability in complex settings, we further evaluate our method on real-world datasets that deviate substantially from the training distribution in terms of mechanisms, scale, and sparsity. 

We adopt CIPCaD-Bench~\cite{menegozzo2022bench} for causal discovery in real-world industrial systems. 
Within this benchmark, \textbf{Ultra-processed Food} models causality in a food production pipeline, containing 17 variables and 83 edges, and can be approximately regarded as a dense ER(5) graph.
This is particularly challenging as the training data reaches a maximum density of only ER(3).
\textbf{Tennessee Eastman Process} originates from a chemical process system and consists of 33 variables with 32 edges, corresponding to a sparse ER(1) graph.
Building on this benchmark, we derive a weak knowledge encoded as $\{-1,0\}$ from the industrial pipeline topology, capturing the constraint that downstream components cannot causally affect upstream ones. 
We illustrate it as \textit{(with topo-priors)} in Tab. \ref{industry}.
In addition, we include the \textbf{Sachs} dataset~\cite{sachs2005causal}, a widely used benchmark in biological causal discovery. 
This dataset captures protein signals and contains 11 variables with 18 edges, which can be viewed as an ER(1.5) graph; here we assume sufficient prior knowledge encoded as $\{-1,0,1\}$.
This is also challenging as the model has never encountered graphs of such minimal scale during training.
Furthermore, all three datasets feature complex, out-of-distribution mechanisms beyond the training ones.

\begin{figure}[t]
  \centering
  \includegraphics[width=0.8\linewidth]{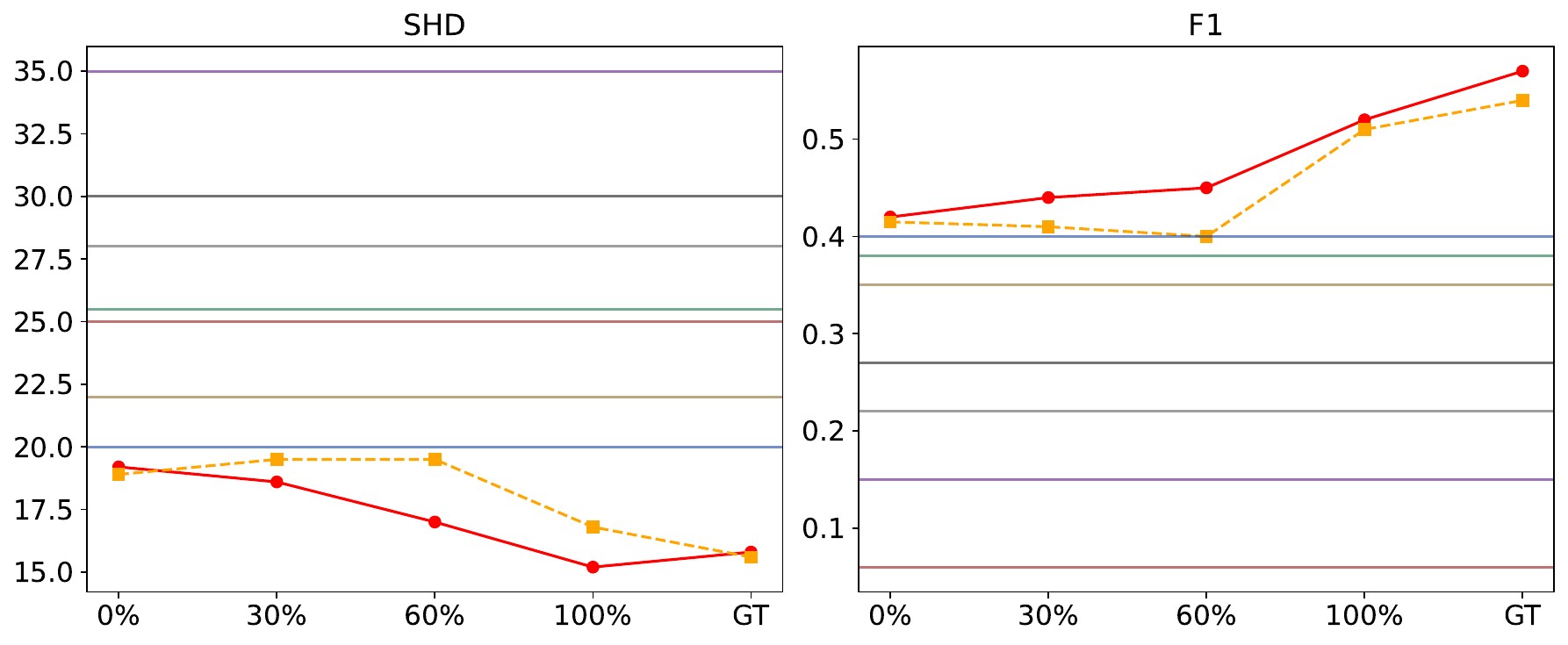}  
  \includegraphics[width=0.7\linewidth]{material/bar.pdf}
  \caption{Sachs Experiment}
  \label{fig:sachs}
\end{figure}

As shown in Fig.~\ref{fig:sachs} and Tab.~\ref{industry}, our method maintains competitive performance across three real-world datasets with diverse scales and mechanisms, even when positive knowledge is absent. 
The best and second-best results are highlighted in bold and underlined, respectively.
In particular, these benchmarks involve substantial shifts in the mechanism, density and graph size relative to the training data, demonstrating the practical utility of our approach in realistic scenarios. 
This resilience stems from the exposure to diverse causal mechanisms during pretraining, which facilitates robust out-of-distribution generalization. 
Furthermore, the synergy between mixed-scale training and prior knowledge integration enables the model to perform scale-aware reasoning, accurately capturing causal structures across heterogeneous graph sizes.

\begin{table*}[t]
\caption{Industry Experiments}
\label{industry}
\centering
\begin{tabular}{ccccc}
\hline
  & \multicolumn{2}{c}{\textbf{Food}} & \multicolumn{2}{c}{\textbf{TEP}} \\
 & \textbf{SHD} & \textbf{F1} & \textbf{SHD} & \textbf{F1} \\ \hline
\begin{tabular}[c]{@{}c@{}}\textbf{Kode}\\ (with topo-priors)\end{tabular} & { \textit{\textbf{80.1}}}       & {\textit{\textbf{0.40}}}       & { \textit{\textbf{31.26}}}       & {\textit{\textbf{0.44}}}      \\
\begin{tabular}[c]{@{}c@{}}\textbf{Kode}\\ (without priors)\end{tabular} &  88.35       & {\ul 0.35}       & {\ul 35.47  }     & {\ul 0.38  }    \\
\textbf{LiNGAM}      & 89.66           & 0.08        & 48.82           & 0.14        \\
\textbf{PC}          & 95.23           & 0.19        & 68.57           & 0.14        \\
\textbf{GES}         & 97.24           & 0.32        & 124.35          & 0.09        \\
\textbf{CSIvA} & 105.3           & 0.17        & 49.77           & 0.22        \\
\textbf{AVICI} & {\ul 87.8}           & 0.22        & 52.61           & 0.28        \\
\textbf{BCNP} & 92.38           & 0.28        & { 35.58}           & 0.31        \\\hline

\end{tabular}

\end{table*}

\section{Conclusion}
In this work, we propose the first knowledge-informed pretrained model for causal discovery, introducing a new paradigm that integrates data and coarse-grained knowledge.
We develop a tailored model architecture, with a corresponding training dataset and curriculum learning strategy to support it, and extensive experiments demonstrate its consistent superiority across diverse settings.
Together, these results highlight the effectiveness and generality of our approach for causal discovery under varying mechanisms, scales, and levels of prior knowledge, both i.d. and o.o.d. .

\clearpage

\bibliographystyle{unsrt}
\bibliography{ref}
\newpage
\end{document}